\pdfoutput=1
\documentclass[letterpaper, 10 pt, conference]{ieeeconf}

\IEEEoverridecommandlockouts                              
\usepackage{cite}
\usepackage{amsmath,amssymb,amsfonts}
\usepackage{algorithmic}
\usepackage[ruled,vlined,linesnumbered]{algorithm2e}
\usepackage{graphicx}
\usepackage{textcomp}
\usepackage{xcolor}
\usepackage{colortbl}   
\usepackage{nicefrac}
\usepackage{comment}
\usepackage{wrapfig}
\usepackage{multirow}
\usepackage{makecell}
\usepackage{subfig}
\usepackage[outercaption]{sidecap}
\usepackage{caption}
\usepackage{bm,times}
\usepackage{booktabs}
\usepackage{cuted}
\usepackage{capt-of} 
\usepackage{amsthm}

\usepackage{pifont}

\usepackage{siunitx}
\usepackage{tikz}
\definecolor{tnom}{rgb}{0,0,0}            
\definecolor{tdet}{rgb}{0.22,0.71,1.00}   
\definecolor{tblk}{rgb}{0.86,0.31,0.31}   
\definecolor{tcsd}{rgb}{0.16,0.42,0.84}   
\definecolor{mstart}{rgb}{0.13,0.55,0.28}  
\definecolor{mend}{rgb}{0.80,0.15,0.15}    
\definecolor{tink}{rgb}{0.16,0.18,0.22}

\definecolor{oursband}{rgb}{0.92,0.95,0.99}

\usetikzlibrary{arrows.meta, positioning, calc, fit, backgrounds, patterns}
\usepackage{url}
\usepackage{xurl}
\usepackage[hidelinks]{hyperref}

\def\BibTeX{{\rm B\kern-.05em{\sc i\kern-.025em b}\kern-.08em
    T\kern-.1667em\lower.7ex\hbox{E}\kern-.125emX}}
\begin{document}

\title{Close\&Safe: Maximal Task Preservation for Redundant Manipulators via C-Space Boundary Detour}

\author{
Abhigyan Singh\textsuperscript{1,2},
Manthan Goyal\textsuperscript{1},
Fares Abu-Dakka\textsuperscript{3},
and Ravi Prakash\textsuperscript{1}
\thanks{
\textsuperscript{1} Indian Institute of Science (IISc), Bengaluru, India.
\textsuperscript{2} Indian Institute of Technology Ropar, Rupnagar, India.
\textsuperscript{3} New York University Abu Dhabi, Abu Dhabi, UAE.
}
\thanks{This work has been submitted to the IEEE for possible publication. Copyright may be transferred without notice, after which this version may no longer be accessible.}
}

\maketitle

\setlength{\stripsep}{0pt}
\setlength{\abovecaptionskip}{2pt}

\begin{strip}
\vspace{-15mm}
\centering
\includegraphics[width=\textwidth]{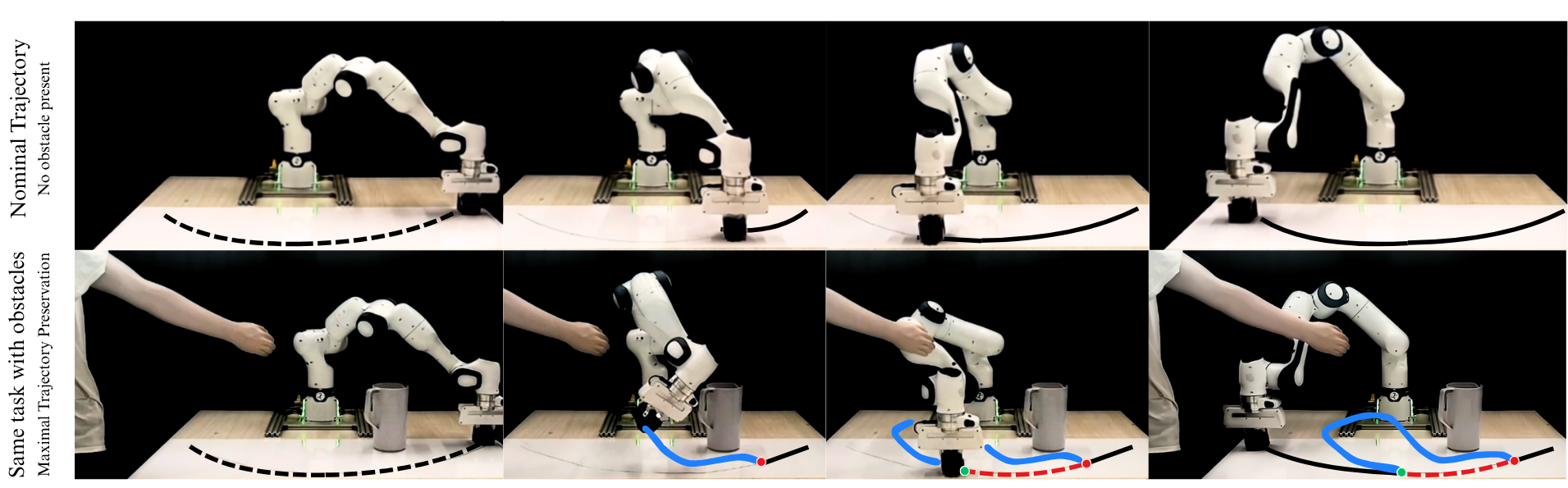}
\vspace{1.2ex}
{\footnotesize\bfseries Visualization of the configuration-space obstacle the jug induces}\\[0.4ex]
\resizebox{\textwidth}{!}{
\begin{tikzpicture}[
  font=\scriptsize,
  cpt/.style={font=\scriptsize, text=black!60, anchor=north, align=center},
]
\newcommand{\framed}[5]{%
  \begin{scope}
    \clip[rounded corners=2.2pt] ({#1},{#2}) rectangle ({#1+#3},{#2+#4});
    \node[inner sep=0pt, outer sep=0pt] at ({#1+#3/2},{#2+#4/2})
      {\includegraphics[width=#3 cm]{#5}};
  \end{scope}
  \draw[rounded corners=2.2pt, draw=black!30, line width=0.6pt]
    ({#1},{#2}) rectangle ({#1+#3},{#2+#4});}

\newcommand{\keyrule}[1]{\tikz[baseline=-0.55ex]{\draw[line width=1.2pt,#1](0,0)--(0.48,0);}}
\newcommand{\keypatch}{\tikz[baseline=-0.55ex]{\fill[black!16,draw=black!45,line width=0.5pt](0,-0.09) rectangle (0.38,0.13);}}
\newcommand{\keydot}[1]{\tikz[baseline=-0.55ex]{\fill[#1,draw=white,line width=0.4pt](0,0) circle (1.5pt);}}

\def\pw{4.14}\def\ph{2.84}\def\pg{0.35}\def\by{0.34}

\framed{0}{\by}{\pw}{\ph}{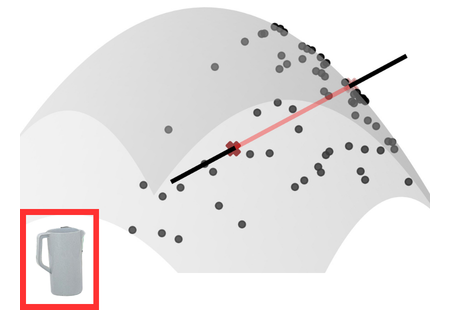}
\framed{\pw+\pg}{\by}{\pw}{\ph}{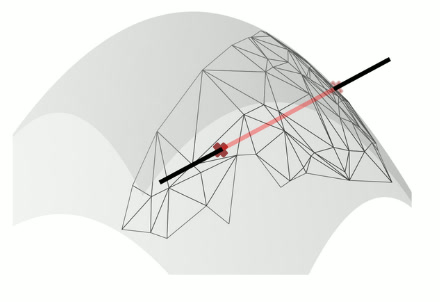}
\framed{2*(\pw+\pg)}{\by}{\pw}{\ph}{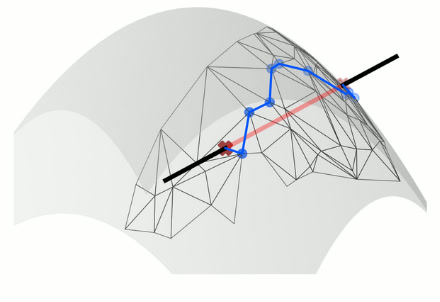}
\framed{3*(\pw+\pg)}{\by}{\pw}{\ph}{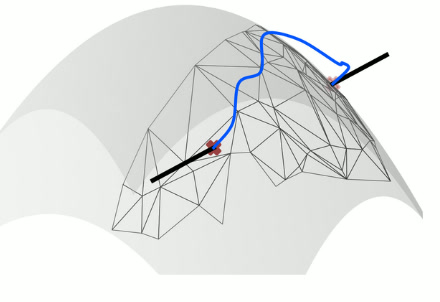}

\foreach \i/\ax/\ay/\bx/\byy in {0/2.13/1.39/3.25/2.08,
                                 1/2.13/1.39/3.25/2.08,
                                 2/2.13/1.39/3.25/2.08,
                                 3/2.00/1.47/3.13/2.12} {
  \node[circle, fill=mend, draw=white, line width=0.5pt, inner sep=1.5pt]
    at ({\i*(\pw+\pg)+\ax},{\by+\ay}) {};
  \node[circle, fill=mstart, draw=white, line width=0.5pt, inner sep=1.5pt]
    at ({\i*(\pw+\pg)+\bx},{\by+\byy}) {};}

\foreach \i/\t in {0/{Corridor sampling}, 1/{Boundary roadmap},
                   2/{Least-deviation search}, 3/{Outward projection}}
  \node[cpt] at ({\i*(\pw+\pg)+\pw/2},{\by-0.06}) {\t};

\foreach \i in {0,1,2}
  \draw[-{Latex[length=1.9mm,width=1.5mm]}, draw=black!50, line width=0.7pt]
    ({\i*(\pw+\pg)+\pw+0.06},{\by+\ph/2}) -- ({(\i+1)*(\pw+\pg)-0.06},{\by+\ph/2});

\node[anchor=south, font=\scriptsize, inner sep=0pt]
  at ({(4*\pw+3*\pg)/2},{\by+\ph+0.22}) {%
    \keyrule{black}~\textcolor{black!70}{nominal}\hspace{1.0em}
    \keyrule{tblk}~\textcolor{tblk}{blocked segment}\hspace{1.0em}
    \keyrule{tcsd}~\textcolor{tcsd}{detour}\hspace{1.0em}
    \keypatch~\textcolor{black!60}{C-obstacle}\hspace{1.0em}
    \keydot{mend}~\textcolor{black!70}{detour start}\hspace{1.0em}
    \keydot{mstart}~\textcolor{black!70}{detour end}};
\end{tikzpicture}}

\captionof{figure}{\textbf{Maximal task preservation under an obstacle unknown at planning time.}
\textbf{First row:} the nominal cleaning path, executed unmodified when nothing obstructs it.
\textbf{Second row:} the same task with a jug and human hand in between that was absent from the planning model. The
executed trajectory follows the nominal exactly outside the blocked segment, departs from it
only where no collision-free configuration tracks the task, and rejoins at the first feasible
configuration. \textbf{Bottom:} the four stages of the detour, shown on the
configuration-space obstacle (grey region) induced by the jug. Red marks the configuration
where the nominal can no longer be followed, green where it resumes. These are the same two
configurations in both rows. Each panel is a projection; the obstacle boundary is a
hypersurface in the seven-dimensional configuration space. \textit{\textbf{*}Please note that the C-space depiction is for visualization purposes only as the original C-space is seven dimensional.}}
\label{fig:teaser}
\end{strip}
\begin{abstract}
Contact-rich manipulation such as wiping, polishing and surface inspection
specifies the task as a continuous end-effector trajectory rather than a goal
pose. When an obstacle absent from the planning model intersects it,
conventional planners return a collision-free path that may differ from the
nominal throughout, discarding trajectory that was never obstructed. We present \textsc{Close\&Safe} (Close to the intended task, Safe from obstacles), which decouples tracking from avoidance: the nominal is followed exactly
wherever it remains feasible, and avoidance acts only on the interval where it
is not, so tracking is hindered as little as the obstacle allows. On that interval we sample the configuration-space obstacle boundary
within a corridor around the nominal, connect the samples into a roadmap, and
search it for the detour that deviates least before rejoining the nominal at
the earliest feasible configuration. We evaluate across diverse simulated obstacle configurations, physical Franka experiments, and ablations of key algorithmic components. The results demonstrate collision-free trajectory repair with localized task modification, while recovering $52.4\text{--}78\%$ in simulation and on hardware.
\end{abstract}

\section{Introduction}
\label{sec:intro}

\begin{table*}[t]
\centering
\caption{Requirements imposed by a prescribed-trajectory manipulation task,
and the extent to which each family of obstacle-avoidance methods satisfies
them. Families are distinguished by decision horizon rather than solver type:
per-instant IK and trajectory optimization are both optimization-based, and
differ in whether avoidance is resolved at a single control instant or across
the whole horizon. $\checkmark$ = satisfied by construction;
$\circ$ = satisfied only conditionally, or traded off against a competing
objective; --- = not addressed by the formulation.}
\label{tab:requirements}
\resizebox{\textwidth}{!}{%
\begin{tabular}{lccccc}
\toprule
\textbf{Task Requirement}
& \begin{tabular}{@{}c@{}}\textbf{Trajectory Opt.}\\ \cite{chomp,Trajopt,stomp,curobo,mpc_manipulator}\end{tabular}
& \begin{tabular}{@{}c@{}}\textbf{Sampling-based}\\ \cite{RRT,PRM,cbirrt,compnetx}\end{tabular}
& \begin{tabular}{@{}c@{}}\textbf{Reactive Repulsion}\\ \cite{apf,ds_modulation,mppi}\end{tabular}
& \begin{tabular}{@{}c@{}}\textbf{Per-instant IK}\\ \cite{collisionik,taskcbf,neural_implicit_sdf,redundancy1,task_priority}\end{tabular}
& \begin{tabular}{@{}c@{}}\textbf{Ours}\\ \phantom{[0]}\end{tabular} \\
\midrule

Exact tracking where collision-free
& $\circ$ & --- & $\circ$ & $\circ$ & $\checkmark$ \\

Deviation confined to blocked segment
& --- & --- & $\checkmark$ & $\checkmark$ & $\checkmark$ \\

Minimal task-space deviation
& $\circ$ & --- & --- & $\circ$ & $\checkmark$ \\

Escapes concave C-space obstacles
& $\circ$ & $\checkmark$ & --- & --- & $\checkmark$ \\

Fallback when redundancy exhausted
& $\circ$ & $\checkmark$ & --- & --- & $\checkmark$ \\

\bottomrule
\end{tabular}%
}
\end{table*}

Robotic manipulators are increasingly used for tasks in which the motion of
the end-effector is not a means of reaching a goal but the specification of
the task itself. In wiping, polishing, welding, and surface inspection a
nominal Cartesian trajectory encodes spatial coverage and contact angle, and
departing from it has task-level consequences: a rerouted wiping motion
leaves an uncleaned region, a displaced weld path produces a seam defect, a
perturbed scan yields gaps in coverage. For such a task, producing some
collision-free path between two task-space poses is not a solution. The
nominal trajectory is a specification to be preserved, not an initial guess a
planner may reshape.

The difficulty arises when an obstacle is absent from the model used to
generate that trajectory. In Fig.~\ref{fig:teaser} a wiping trajectory is
planned across a table and a jug and human(mannequin), unmodeled at planning time, collides in part
of the swept region. The obstacle invalidates a short segment of the motion
and the remainder stays executable. Four requirements follow. The executed
trajectory should track the nominal wherever it remains feasible, depart from
it only over the obstructed segment, minimize the task-space deviation
incurred during that departure, and rejoin the nominal at the earliest
configuration from which it can be resumed. We refer to this objective as
\emph{maximal task preservation}.

Existing methods trade these requirements against one another rather than
ordering them, and the horizon over which avoidance is decided governs how.
Trajectory optimization~\cite{chomp, stomp, Trajopt, curobo,
mpc_manipulator} and sampling-based planning~\cite{RRT, PRM, cbirrt,
compnetx} decide over the whole trajectory and are free to reshape all of it.
Reactive repulsion~\cite{apf, ds_modulation, neural_implicit_sdf, mppi} and
per-instant inverse kinematics~\cite{collisionik, taskcbf} decide at one
control instant and cannot look ahead, so a deviation is sized by a gain or a
greedy step rather than by the segment to be covered. Null-space redundancy
resolution~\cite{redundancy1, task_priority} is the exception: it tracks the
nominal exactly while a collision-free redundant configuration exists, which
is why we adopt it as the first level of our hierarchy, and offers nothing
once that manifold is blocked. Table~\ref{tab:requirements} records which
requirements each family meets; Sec.~\ref{sec:related} treats them in
detail.

We propose \textsc{Close\&Safe} (Close to the intended task, Safe from obstacles), which orders these requirements instead of weighting them. Interference from an obstacle is classified as \emph{reducible}, meaning null-space motion can absorb it with no task deviation, or \emph{irreducible}, meaning no configuration satisfies the task and avoids collision at once. The classification is a feasibility test rather than a comparison against a margin, so it decides when the task may be modified without a quantity to tune. Where interference is irreducible, we take the behavioral principle of Bug algorithms~\cite{bug1, bug2}: follow the reference, detour along the obstacle boundary, and rejoin at the first opportunity, but replace greedy tracing with an explicit road-map of the configuration-space obstacle boundary, built in a corridor around the blocked segment and searched. The circumnavigation direction is selected by comparing candidate routes across the blocked segment, rather than by a local gradient, extending the boundary-following principle beyond planar settings. Restricting exploration to a thin boundary shell also avoids sampling unnecessary free-space volume, reducing the computation required for the detour search. Fig.~\ref{fig:teaser} shows the behavior and the four stages that produce it. Our contributions are:

\begin{itemize}
    \item A minimum-deviation detour confined to the blocked segment, found
    by searching a roadmap of the configuration-space obstacle boundary built
    in a corridor around it. Concave and multiply-connected geometry needs no
    special treatment, and segments never in collision are never discarded.

    \item An evaluation against four baselines on four simulated Franka
    scenarios, scored by coverage, the fraction of the nominal path executed
    within tolerance. Among methods collision-free throughout, ours
    attains the highest preservation in all three cleaning cases, averaging
    $59.5\%$ against $45.6\%$, at a clearance held within a $4.5\,$mm band.
    Absolute values are bounded by the scenarios, in which part of the path is
    unreachable for any method.

    \item Hardware validation on two tasks with demonstration-derived
    nominal: cleaning with a jug and human(mannequin) in the path and scanning interrupted by a human(mannequin) hand posture.
\end{itemize}

\section{Related Work}
\label{sec:related}

We group prior work by the horizon over which obstacle avoidance is decided,
since that is what determines whether an existing trajectory can be kept, and
treat separately the methods that deform a given path rather than generate a
new one.

\subsection{Whole-Horizon Methods}
Trajectory optimization treats the entire path as the decision variable and
minimizes a weighted objective over it, by covariant gradient
descent~\cite{chomp, chomp_ijrr}, sampled rollouts~\cite{stomp}, sequential
convex programming~\cite{Trajopt}, or GPU-parallel solvers fast enough for a
receding horizon~\cite{curobo, mpc_manipulator}. A deviation penalty makes
preservation one term among several, weighted against clearance and
smoothness and retuned per scene, while imposing the nominal as a hard
constraint leaves the program infeasible exactly when no collision-free
configuration satisfies the task. Either way the decision variable is the
whole trajectory, so one localized obstacle can reshape motion that was never
in collision.

Sampling-based planners build a graph of collision-free configurations and
search it for connectivity~\cite{PRM, RRT}. They are not defeated by concave
or multiply-connected geometry, the advantage this paper borrows, and
constrained variants restrict the search to a task manifold by projection onto
Task Space Regions~\cite{cbirrt} or by learned generators~\cite{compnetx}.
Their objective is feasibility between endpoints rather than similarity to a
reference, so a planner invoked on a partially obstructed trajectory may
return a path differing from the nominal throughout, with no criterion
bounding the drift.

\subsection{Instant-Horizon Methods}
Reactive methods superpose an attractive term pulling the robot toward the
reference with repulsive terms from nearby obstacles, as potential
fields~\cite{apf} or modulation of a stable dynamical
system~\cite{ds_modulation}; learned distance fields make the proximity
queries cheap for articulated bodies~\cite{neural_implicit_sdf, cdf_rss} and
can drive sampling-based model-predictive control~\cite{mppi}. The
modification is confined to the obstacle's vicinity, the right granularity,
but the reasoning is memoryless. Without a model of configuration-space connectivity, these controllers stall in local minima around concave geometry, with deviation determined by the gains. Repulsion acts at a standoff distance, so in contact-rich tasks, the surface the tool must track itself repels, requiring a specialized formulation.

Per-instant inverse kinematics resolves avoidance at each control step, by
weighting pose against clearance in a nonlinear program~\cite{collisionik} or
enforcing safety as a constraint while tracking the task~\cite{taskcbf}. For
redundant manipulators the avoidance motion can be projected into the
Jacobian null space and absorbed without perturbing the end-effector at
all~\cite{redundancy1, task_priority}, and global maps over the redundant
degree of freedom~\cite{swivel_plus_end-eff} or dynamic programming over a
discretized self-motion manifold~\cite{dp_criteria, DP_ros, dp_prescribed}
resolve it over a whole prescribed path rather than greedily. This family
preserves the task exactly and is adopted here as the first level of our
hierarchy, but all of it assumes the task remains satisfiable: once an
obstacle blocks the entire self-motion manifold, no redundant configuration is
collision-free and the problem as posed has no solution.

\subsection{Path Deformation and Boundary Following}
A third line takes an existing path as the object to be modified, the setting
of this paper. Elastic bands~\cite{elastic_khatib, elastic_collision} model
the path as a deformable curve under internal contraction and external
repulsion, so it adapts to the scene while staying near its original shape.
The deformation is driven by the same local forces as the reactive methods
above: a concave obstacle traps the band, the deviation is set by the balance
of the two force terms rather than by what the obstacle requires, the contact
surface repels, and the band deforms whenever an obstacle is in range, with no
test distinguishing a segment the robot can still execute from one it cannot.

Bug algorithms~\cite{bug1, bug2} provide the behavior we adopt where the task
is not satisfiable: follow the reference, detour along the obstacle boundary,
rejoin at the first opportunity. They were formulated for a point in the plane
with local sensing, tracing the boundary greedily in a direction chosen
without knowledge of the alternative. Neither property survives the move to
seven dimensions, where the boundary is not a curve and cannot be followed by
contact. We retain the behavior and replace the tracing with an explicit
roadmap of the configuration-space obstacle boundary, built within a corridor
around the blocked segment and searched, so the direction of circumnavigation
follows from comparison across the segment.

\section{Preliminaries and Problem Formulation}
\label{sec:prelim}

\subsection{Robot Model and Redundancy Resolution}
\label{subsec:prelim_redundancy}
Consider an $n$-DOF redundant manipulator with configuration space
$\mathcal{Q} \subset \mathbb{R}^n$, abbreviated C-space, and forward
kinematics $f_{\text{FK}}: \mathcal{Q} \to \text{SE}(3)$. A nominal task is a
Cartesian curve $\mathcal{C}_{\text{nom}} = \{ x_{\text{nom}}(s) \mid s \in
[0,1] \}$ traversed over $t \in [0,T]$, with static obstacles $\mathcal{O}
\subset \mathbb{R}^3$. Approximating the robot by $B$ bounding spheres gives
a differentiable minimum signed distance $d(q,\mathcal{O})$~\cite{chomp}, and
with joint limits $\mathcal{Q}_{\text{feas}}$ and clearance margin
$d_{\text{safe}} > 0$ the collision-free C-space is
$\mathcal{Q}_{\text{free}} = \{ q \mid d(q,\mathcal{O}) \ge d_{\text{safe}}
\}$.

The null space is parameterized by an elbow swivel angle
$\phi$~\cite{kinematic_analysis}. Candidates $q_{i,k} =
f_{\text{IK}}(x_{\text{nom}}(t_i), \phi_{i,k})$ are generated along the
nominal, and dynamic programming over them~\cite{dp_prescribed, DP_ros}, with
$+\infty$ node costs at collisions and transition costs penalizing joint and
swivel displacement, yields a task-preserving realization
$Q_{\text{swivel}}$. This is prior work and forms the first layer.

\subsection{Admissibility and Interference}
The self-motion manifold $\mathcal{M}_{\text{task}}(s) = \{ q \mid
f_{\text{FK}}(q) = x_{\text{nom}}(s) \}$ collects the configurations
satisfying the pose exactly, and those also feasible and collision-free are
\begin{equation}
\label{eq:admissible}
\mathcal{S}(s) = \mathcal{M}_{\text{task}}(s) \cap \mathcal{Q}_{\text{feas}} \cap \mathcal{Q}_{\text{free}}.
\end{equation}
Interference is \emph{reducible} where $\mathcal{S}(s) \neq \emptyset$, since
null-space reconfiguration absorbs the obstacle without perturbing the task,
and \emph{irreducible} where $\mathcal{S}(s) = \emptyset$. The irreducible
domain $\mathcal{S}_{\text{infeas}} = \{ s \mid \mathcal{S}(s) = \emptyset
\}$ is the only part of the task the method may modify, and deciding between
the cases is an emptiness test on \eqref{eq:admissible}, so the condition
carries no threshold to tune.

\subsection{Task-Preservation Metric}
\label{sssec:tp_metric}

We measure task preservation as the fraction of the nominal trajectory for
which the executed trajectory contains a point satisfying prescribed position
and orientation tolerances. The nominal trajectory is sampled at $M$
checkpoints uniformly in arc length. A checkpoint $s_j$ is preserved if
there exists an executed point $u$ such that
\begin{equation}
\begin{aligned}
\left\|p_{\text{exec}}(u)-p_{\text{nom}}(s_j)\right\|_2
&\leq \epsilon_{\text{pos}},\\
2\arccos\left(
\left|\mathbf q_{\text{exec}}(u)\cdot
\mathbf q_{\text{nom}}(s_j)\right|
\right)
&\leq \epsilon_{\text{ori}}
\end{aligned}
\end{equation}
Thus,
\begin{equation}
\label{eq:tp}
\mathrm{TP}(Q)
=
\frac{1}{M}\sum_{j=1}^{M}
\mathbb I_{\text{pres}}(s_j)\times100\%,
\end{equation}
where $\mathbb I_{\text{pres}}(s_j)=1$ when such a point exists.
Since checkpoints are uniform in arc length, $\mathrm{TP}$ directly measures
the fraction of the nominal stroke preserved.

\subsection{Problem Statement}
Given $\mathcal{C}_{\text{nom}}$, $\mathcal{O}$, and $d_{\text{safe}}$, we seek
a trajectory $Q$ optimized in three lexicographically ordered levels rather
than as a weighted sum. \emph{Feasibility} requires $q_k\in
\mathcal{Q}_{\text{feas}}$ and $d(q_k,\mathcal{O})\ge d_{\text{safe}}$
everywhere, including between waypoints, while satisfying the velocity,
acceleration, and jerk limits. \emph{Task preservation} then maximizes
$\mathrm{TP}(Q)$ among feasible trajectories, restricting deviation from
$\mathcal{C}_{\text{nom}}$ to $\mathcal{S}_{\text{infeas}}$.
\emph{Minimal detour} then minimizes joint-space excursion and task-space
deviation over the modified interval. This ordering ensures that deviation is introduced
only when required for feasibility and is minimized thereafter.

\section{Methodology}
\label{sec:method}
\textsc{Close\&Safe} attains the objective of Sec.~\ref{sec:prelim} by
minimal intervention, in two layers that correspond to the two cases the
classification separates:

\begin{enumerate}
    \item \textbf{Null-space redundancy resolution.} Where interference is
    reducible, redundancy is resolved along the exact task path by the
    swivel-angle dynamic programme of
    Sec.~\ref{subsec:prelim_redundancy}~\cite{dp_prescribed, DP_ros,
    swivel_plus_end-eff}, yielding a candidate trajectory
    $Q_{\text{swivel}}$ with zero task deviation.

    \item \textbf{Localized C-space boundary detour.} Where it is
    irreducible, the framework isolates the maximal contiguous interval of
    $\mathcal{S}_{\text{infeas}}$, referred to throughout as the
    \emph{blocked segment}, and repairs that interval alone.
\end{enumerate}

\begin{algorithm}[ht]
\caption{Minimal-Intervention Localized C-Space Repair}
\label{alg:cspace_repair}
{\small
\begin{algorithmic}[1]
\REQUIRE $Q_{\text{swivel}}$, obstacle $\mathcal{O}$, $d_{\text{safe}}$
\STATE \textbf{Identify} intervals $\mathcal{I}_{\text{repair}} = [i_{\text{entry}}, i_{\text{exit}}]$ in $Q_{\text{swivel}}$
\FOR{\textbf{each} merged interval $\mathcal{I}_{\text{repair}}$}
    \STATE Define local corridor bounds $q_{\text{min}}^c$, $q_{\text{max}}^c$
    \STATE Generate Sobol samples in bounded corridor
    \STATE $\mathcal{P}_{\text{boundary}} \leftarrow \{q : d(q,\mathcal{O}) \leq \epsilon_{\text{shell}}\}$
    \STATE Build $k$-NN roadmap on $\mathcal{P}_{\text{boundary}}$
    \STATE $\Pi_{\text{detour}} \leftarrow$ A$^*$ from $q_{\text{entry}}$ to $q_{\text{exit}}$
    \FOR{\textbf{each} $q \in \Pi_{\text{detour}}$ with $d(q) < d_{\text{safe}}$}
        \STATE Project $q$ outward along $\nabla_q d(q, \mathcal{O})$
    \ENDFOR
    \STATE Validate edges via recursive adaptive bisection
    \STATE $Q_{\text{candidate}}[\mathcal{I}_{\text{repair}}] \leftarrow \Pi_{\text{detour}}$
\ENDFOR
\STATE Apply PCHIP and minimum-jerk time parameterization
\RETURN $Q_{\text{candidate}}$
\end{algorithmic}}
\end{algorithm}

The test $\mathcal{S}(s) = \emptyset$ is evaluated at each waypoint, so layer
two may be invoked more than once along a single trajectory and is not
invoked at all where redundancy suffices.

\subsection{Minimal-Intervention Localized C-Space Repair}
\label{subsec:detour_novelty}

When an obstacle invalidates the null-space manifold over a contiguous
sequence of waypoints, $Q_{\text{swivel}}$ contains collisions. Rather than
replanning globally, we extract and repair only the invalid segments.

\subsubsection{Irreducible Repair Interval Localization}

The two configurations this step returns are the marks carried
through every panel of Fig.~\ref{fig:teaser}. The signed-distance profile of
$Q_{\text{swivel}}$ is scanned to identify contiguous collision intervals. For each interval, safe entry
($i_{\text{entry}}$) and exit ($i_{\text{exit}}$) indices are identified
where the clearance meets $d_{\text{safe}}$, giving the repair interval
$\mathcal{I}_{\text{repair}}=[i_{\text{entry}},i_{\text{exit}}]$.

\subsubsection{Localized Corridor and Boundary-Focused Sampling}

To find a compact detour, exploration is restricted to a bounded local corridor
in C-space. Anisotropic bounds $q_{\text{min}}^c$ and
$q_{\text{max}}^c$ are defined around $q_{\text{entry}}$ and
$q_{\text{exit}}$ using joint-specific margins, shown in the first panel in the lower row of
Fig.~\ref{fig:teaser}.

Within this corridor, low-discrepancy Sobol samples are
generated~\cite{sobol_sampling}. To bias the search toward minimal
excursions that tightly circumnavigate the obstacle, configurations are
filtered to retain only those lying within a thin shell around the C-space
obstacle boundary:
\begin{equation}
\begin{split}
\mathcal{P}_{\text{boundary}} &= \Big\{ q\in\mathcal{B}_{\text{corridor}} \mid \\
&\quad 0\leq d(q,\mathcal{O})\leq\epsilon_{\text{shell}} \Big\}.
\end{split}
\end{equation}

\subsubsection{Roadmap construction and search}
Samples outside the joint limits are discarded before connection, so
every road-map node is admissible and joint limits are enforced by
construction. The retained samples are connected into a directed
$k$-nearest-neighbor roadmap over
$\mathcal{P}_{\text{boundary}}$~\cite{PRM}, drawn as the mesh in the
second panel of Fig.~\ref{fig:teaser}. Edges are not collision-checked at
build time: every node is later projected outward to clearance
$d_{\text{safe}}$ and every retained edge is validated by bisection
(Sec.~\ref{subsubsec:projection}), so the check is deferred rather than
omitted. Connections are additionally capped at a maximum radius
$r_{\max}$, which keeps edges short enough that they cannot span the
obstacle from one side to the other. It is searched by
A$^*$~\cite{Astar} from $q_{\text{entry}}$ to $q_{\text{exit}}$ under the
edge cost
\begin{equation}
\begin{split}
w(q_a,q_b) = {} & \underbrace{\lVert q_b - q_a \rVert_W}_{\text{C-space length}}
+ \lambda_x \underbrace{\lVert p(q_b) - p_{\text{ref}}(q_b) \rVert_2}_{\text{task deviation}} \\
& + \lambda_d \max\!\big(0,\, d_{\text{safe}} - d(q_b)\big),
\end{split}
\label{eq:astar_cost}
\end{equation}

where $p(\cdot)$ is the end-effector position under forward kinematics,
$p_{\text{ref}}(q_b)$ is the nearest point on the nominal task path, and
the last term penalizes clearance below $d_{\text{safe}}$. The heuristic
is the w, weighted distance to $q_{\text{exit}}$, which is admissible
since the two penalty terms are non-negative. The search therefore
returns the detour of least joint displacement that also stays close to
the nominal task path, given the highlighted traverse in blue in the
third panel.

We keep both the C-space length and the task-deviation term by design
rather than necessity: the two are not equivalent, since a short joint
motion may still swing the end-effector far from the nominal path. Using
$\lambda_x$ to weight them lets the balance be set per task, raising
$\lambda_x$ when the end-effector path itself carries the task
semantics, and lowering it towards $\lambda_x \!=\! 0$ when the task is
already relinquished in the blocked interval, and the shortest joint
detour is the preferable one.

Because the route is selected by search over the whole road-map rather
than by tracing the boundary greedily, the direction of
circumnavigation follows from global comparison across the sampled
boundary. Concave and multiply-connected geometry, where the shorter
route is not the locally attractive one, therefore requires no special
handling.

\subsubsection{Outward Clearance Correction and Bisection Validation}
\label{subsubsec:projection}

Because nodes reside near the obstacle
boundary, residual penetrations are resolved via gradient-based projection.
Selected road-map points are displaced outward along the steepest-ascent normal $n(q) = \nabla_q d /
\|\nabla_q d\|_2$ until clearance is achieved:
\begin{equation}
\begin{split}
q &\leftarrow \operatorname{clamp} \Big( q+ \beta \big(d_{\text{safe}}-d(q)+\delta\big)n(q), \\
&\quad q_{\text{min}}, q_{\text{max}} \Big).
\end{split}
\end{equation}
The clamp to $[q_{\text{min}}, q_{\text{max}}]$ makes the projection itself
joint-limit preserving, so clearance is never bought at the cost of a limit
violation. Continuous collision-free connectivity between discrete detour
way-points is guaranteed by recursively bisecting, projecting, and validating
edges.

\subsubsection{Detour Splicing and Continuous Time-Parameterization}

The discrete verified detour $\Pi_{\text{detour}}$, having
satisfied outward clearance projection and recursive bisection validation,
directly replaces
the infeasible interval in the trajectory:
\begin{equation}
Q_{\text{candidate}}[i_{\text{entry}}:i_{\text{exit}}] = \Pi_{\text{detour}}.
\end{equation}
The modified trajectory is arc-length parameterized and fitted per joint with
monotonic Piecewise Cubic Hermite Interpolating Polynomials
(PCHIP)~\cite{pchip}, which guarantees zero overshoot between waypoints and so
cannot reintroduce collisions at the validated knots. A quintic minimum-jerk
phase profile~\cite{quintic} then assigns time along this path, with the total
duration scaled to the smallest value satisfying the joint velocity and
acceleration limits. The resulting trajectory is $C^1$-continuous with bounded,
piecewise-continuous acceleration that vanishes at both endpoints.

\section{Comparison}
\label{sec:comparison}

We compare against four baselines spanning the families of
Sec.~\ref{sec:related}: TrajOpt~\cite{Trajopt} for horizon-wide
trajectory optimization, CollisionIK~\cite{collisionik} and the
quadratic-program controller of Neural-JSDF~\cite{neural_implicit_sdf}
for per-instant inverse kinematics, and the sampling-based
model-predictive controller of the same work for reactive avoidance.
Two choices are deliberately unfavorable to us. We tune every baseline
per scenario using Optuna, 30 trials each, while our own parameters are
held fixed throughout. For the stochastic baseline we report the
best run over seeds and for ours the mean.

All four scenarios run in simulation on a 7-DOF Franka. The nominal
trajectory is a table-wiping stroke of $111.4\,\text{cm}$ for the three
cleaning scenarios and a wall-spraying stroke of $120.4\,\text{cm}$ for
the fourth, sampled at $M=500$ checkpoints and counted as achieved
within $5\,\text{mm}$ and $5^{\circ}$. Case~1 places a vase partway
along the stroke. Case~2 replaces it with a second stationary Franka
reaching across the table. Case~3 combines both. Case~4 sprays a wall
with a second arm standing to one side. The simulation scene along with the generated trajectories is shown in Fig.~\ref{fig:bench} top row.

\begin{table}[!t]
\centering
\caption{Hyperparameters used for the C-space avoidance repair pipeline, with their definitions and default values. These values are held fixed across every experiment unless that parameter is the subject of the ablation.}
\label{tab:hyperparams}
\begin{tabular}{@{} p{0.65\columnwidth} p{0.25\columnwidth} @{}}
\toprule
\textbf{Definition (Symbol)} & \textbf{Value} \\
\midrule
C-obstacle boundary seed samples ($N_s$) & 100,000 \\
Boundary-shell acceptance tolerance ($\epsilon$) & $35$ mm \\
Neighbors per node in the boundary road-map ($k$) & 14 \\
Per-joint anisotropic C-space margin ($\bm{\delta}$) & $[1.45, 0.90, 0.75,$ \newline $1.10,0.40, 0.85, $ \newline $0.40]$ rad \\
Swivel-manifold angular grid resolution ($G$) & 128 \\
\bottomrule
\end{tabular}
\end{table}

\begin{figure*}[t]
    \centering
    \resizebox{\textwidth}{!}{\input{fig_bench}}
    \caption{The four benchmark scenarios. \textbf{Top:} executed
    trajectories rendered in the scene. \textbf{Middle:} minimum clearance to
    the obstacle set along the trajectory; the band below zero is
    penetration, so a curve entering it is inadmissible whatever its task
    preservation. \textbf{Bottom:} the checkpoints each method executed
    within tolerance, placed at their position along the nominal path, so a
    gap is task lost and its position says where it was lost. Lane colors in the bottom row are the colors used for the
    same methods in the rows above: {\color{red}\rule[0.5ex]{8pt}{1.5pt}}~nominal trajectory (red), our proposed {\color{cyan}\rule[0.5ex]{8pt}{1.5pt}}~Close\&Safe (cyan), {\color{green}\rule[0.5ex]{8pt}{1.5pt}}~TrajOpt (green), {\color{orange}\rule[0.5ex]{8pt}{1.5pt}}~CollisionIK (orange), {\color{magenta}\rule[0.5ex]{8pt}{1.5pt}}~Neural-JSDF QP (magenta), and {\color{yellow!80!black}\rule[0.5ex]{8pt}{1.5pt}}~Neural-JSDF MPPI (yellow). The nominal is drawn in red in the upper
    two rows as the reference and is in collision by construction. The hatched red region in the bottom row shows the trajectory violates the collision constraint.
  }
  \label{fig:bench}
\end{figure*}

\begin{table*}[!t]
\centering
\caption{Task preservation ($\mathrm{TP}$, the percentage of the nominal stroke executed within tolerance; higher is better) and minimum signed clearance (positive = separation, negative = penetration) across the four benchmark scenarios. All baselines use hyperparameters obtained by per-scenario constrained Bayesian optimization (Optuna, 30 trials per method-scenario pair), maximizing $\mathrm{TP}$ subject to zero collision and to the Franka per-joint velocity, acceleration and jerk envelope. Entries marked $\dagger$ violate the collision constraint and are not admissible regardless of their $\mathrm{TP}$.}
\label{tab:tp_clearance}
\setlength{\tabcolsep}{5pt}
\begin{tabular}{l cc cc cc cc}
\toprule
& \multicolumn{2}{c}{\textbf{Case 1}: Cleaning (vase)} & \multicolumn{2}{c}{\textbf{Case 2}: Cleaning (arm)} & \multicolumn{2}{c}{\textbf{Case 3}: Cleaning (vase + arm)} & \multicolumn{2}{c}{\textbf{Case 4}: Spray painting} \\
\cmidrule(lr){2-3} \cmidrule(lr){4-5} \cmidrule(lr){6-7} \cmidrule(lr){8-9}
\textbf{Method} & TP (\%) & Clr (mm) & TP (\%) & Clr (mm) & TP (\%) & Clr (mm) & TP (\%) & Clr (mm) \\
\midrule
\textbf{Close\&Safe (Ours)} & \textbf{68.4} & $+17.6$ & \textbf{57.8} & $+13.1$ & \textbf{52.4} & $+14.1$ & 75.0 & $+13.7$ \\
TrajOpt~\cite{Trajopt}                          & 54.4 & $+6.0$  & 35.0$^{\dagger}$ & $-55.4$ & 27.2$^{\dagger}$ & $-107.0$ & \textbf{76.4} & $+5.4$ \\
Neural-JSDF (QP-IK)~\cite{neural_implicit_sdf}  & 19.0$^{\dagger}$ & $-27.5$ & 22.0 & $+9.7$  & 34.6 & $+2.2$  & 75.8 & $+1.1$ \\
Neural-JSDF (MPPI)~\cite{neural_implicit_sdf}   & 35.6 & $+3.5$  & 23.4 & $+21.7$ & 35.6 & $+1.6$  & 38.0 & $+0.5$ \\
CollisionIK~\cite{collisionik}          & 65.0 & $+24.1$ & 22.4 & $+61.3$ & 49.4 & $+31.2$ & 75.2 & $+2.4$ \\
\bottomrule
\end{tabular}
\end{table*}

\subsection{Completion of the nominal path}
The clearest separation is not the preservation percentage but how far
along the nominal path each method gets at all. Ours is the only method
that traverses the full path in all four scenarios. The sampling-based
controller never passes $s \approx 0.39$ in any scenario, including the
two easy ones, and the QP controller does not pass $s \approx 0.35$ in
any cleaning scenario. Both are driven by the same learned joint-space distance field, and
both halt at the first concavity, where every locally available
direction increases cost. Their preservation is a single termination
rather than distributed loss.

\subsection{Easy scenarios}

Case~4 establishes the task ceiling: four of five methods land between $75.0\%$ and $76.4\%$, a spread of only $1.4$ points across entirely different formulations, indicating that the wall fixture makes roughly a quarter of the stroke unreachable regardless of method. Our $75.0\%$ reaches that ceiling, but the clearance column shows why it matters: at $+13.7$\,mm, within our $4.5$\,mm band across all scenarios, we reach it without trading away margin. Because clearance and task preservation are separate, lexicographically ordered layers rather than weighted together, there is no tradeoff between safety and task preservation.

Case~1 is geometrically simple but produces no ceiling, and the spread
is $49.4$ points. One convex obstacle already separates the families.
We reach $68.4\%$ and CollisionIK $65.0\%$, both retaining the stroke
either side of the vase. TrajOpt reaches $54.4\%$ at only
$+6.0\,\text{mm}$, the QP controller collides at $-27.5\,\text{mm}$,
and the sampling-based controller halts at $s=0.36$.

\subsection{Hard scenarios}
In Case~2, the second arm removes the self-motion manifold over a long
interval. We reach $57.8\%$ at $+13.1\,\text{mm}$. TrajOpt reports
$35.0\%$ but penetrates to $-55.4\,\text{mm}$, so its spans are not a
usable result, and both learned-field controllers halt below $s=0.25$.
CollisionIK is the informative failure: it falls to $22.4\%$ while
holding $+61.3\,\text{mm}$, four times our margin in the same scene.
It has not failed to find a safe configuration but has found one far
from the obstacle and stayed there, which is what a per-instant
objective produces with no horizon over which to evaluate the trade.

Case~3 adds concavity, and the narrow strip of table between the vase
and the arm is reachable only through it. We reach $52.4\%$ and CollisionIK $49.4\%$, the only two methods to
recover stroke beyond the obstruction. The top row shows why our
trajectory takes a visibly long excursion: the detour returns the
end-effector to that strip rather than skirting the obstacles at a
distance, and it does so at $+14.1\,\text{mm}$ against CollisionIK's
$+31.2\,\text{mm}$. TrajOpt penetrates to $-107.0\,\text{mm}$.
\subsection{Failure modes}
CollisionIK is a close competitor on the easy scenarios, $70.1\%$
against our $71.7\%$, but on the hard ones the comparison is $35.9\%$
against $55.1\%$: the margin widens from $1.6$ to $19.2$ points.

The clearance row separates three failure modes. Halting at a
concavity gives low preservation at small positive clearance, which is
both learned-field controllers. Over-avoidance gives low preservation
at large positive clearance, which is CollisionIK in Cases~2 and~3.
Violation gives high apparent preservation at negative clearance, which
is TrajOpt in the same cases. None is a tuning failure, since every
baseline was tuned per scenario and still exhibits them.

Our clearance spans $4.5\,\text{mm}$ across all four scenarios, against
$58.9\,\text{mm}$ for CollisionIK and $113.0\,\text{mm}$ for TrajOpt.
Our margin is the specified $d_{\text{safe}}$, produced by outward
projection, not a value emerging from the tuner. Over the three
cleaning scenarios we average $59.5\%$ against $45.6\%$ for the
strongest baseline admissible everywhere.
\section{Hardware Experiments}
\label{sec:hardware}
\begin{figure*}[t!]
\centering
\includegraphics[width=0.65\textwidth]{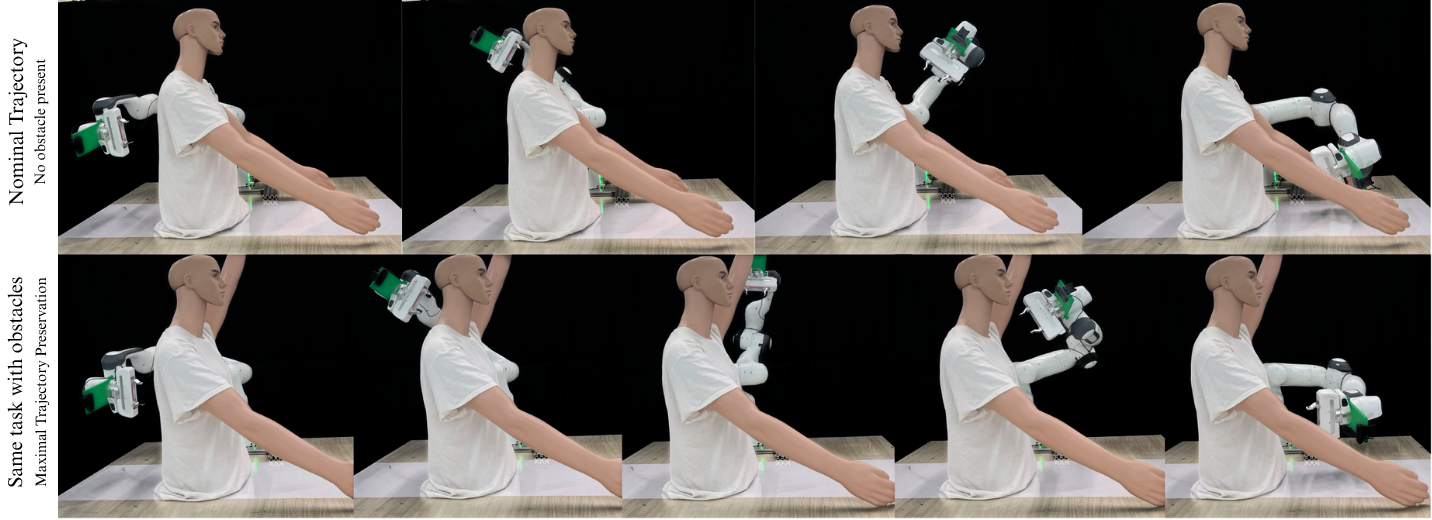}
\caption{Case 2: Scanning on hardware. \textbf{Top:} the nominal sweep, from the back
past the neck to the abdomen, with both arms lowered. \textbf{Bottom:} the
same nominal executed with the left arm raised. The end-effector does not
stall in the pocket between the head and the raised arm but passes around it
and returns to the nominal, retaining $78\%$ of the scan.}
\label{fig:hw_scan}
\end{figure*}

We validate the framework on a physical Franka across two tasks. Each
nominal trajectory is recorded by kinesthetic demonstration and then executed
against an obstacle that was absent when the demonstration was taken.
Obstacles are perceived by a RealSense D435i and are stationary during
execution, so the static obstacle set assumed in Sec.~\ref{sec:prelim} holds.
The human figure is a mannequin, posed once and held fixed.

\textbf{Case 1: Cleaning with two obstructions.} The jug is placed slightly behind
the stroke, and a mannequin arm intrudes from the front over the latter half,
so the free space narrows from opposite sides at different points along the
path. The arm clears both while recovering as much of the stroke as the
remaining space allows. It is the perfect example showing the harmony between null space resolution and minimal boundary detour, providing a task preservation of $73\%$. The jug leads to a detour initially, but the arm lying in the null space only bends the elbow, leaving the end effector unaffected (Fig.~\ref{fig:teaser}).

\textbf{Case 2: Scanning.} The nominal sweeps the end-effector along the mannequin
from the back, past the neck, to the abdomen, with both arms lowered. In the
modified scene the left arm is raised and obstructs the sweep, leaving a
pocket between the head and the raised arm in which a locally guided
controller would stall. The executed motion passes around the raised arm
instead and rejoins the nominal beyond it, retaining $78\%$ of the scan
(Fig.~\ref{fig:hw_scan}). This is the largest excursion of the three tasks
and also the highest preservation, which is not a contradiction: what is lost
is the portion of the nominal that cannot be tracked, not the distance
traveled to clear the obstacle.

\section{Ablation}
\label{sec:ablation}

To isolate the C-space detour stage, we evaluate it on 800 narrow-passage
problems from the MotionBenchMaker suite (defined by start $\bm{q}_0$ and
goal pose) \cite{motionbenchmaker}. With the null-space layer disabled, the initial trajectory is a
straight C-space line. Timings are recorded on an Intel Core~i7 and RTX~4060
GPU. We ablate two design choices: the boundary seed count
(Table~\ref{tab:seed_count}) and sensitivity to the sampling stream
(Table~\ref{tab:seed_variation}). The swivel stage is bypassed throughout so
that the quantity under test is the only source of variation.

\begin{table}[h]
\centering
\caption{Effect of Sobol seed count on repair success and latency.
$^{\ast}$Default configuration.}
\label{tab:seed_count}
\begin{tabular}{@{}rccc@{}}
\toprule
Seeds & Pass (\%) & Median $t$ (ms) & Mean $t$ (ms) \\
\midrule
1,000 & 87.0 & 9.9 & 18.4 \\
2,500 & 92.5 & 7.1 & 12.5 \\
5,000 & 96.2 & 5.8 & 9.8 \\
10,000 & 99.0 & 7.4 & 10.6 \\
25,000 & 99.9 & 15.0 & 18.2 \\
100,000$^{\ast}$ & 100.0 & 30.9 & 36.0 \\
\bottomrule
\end{tabular}
\end{table}

Success saturates at 100\% with $10^5$ seeds (Table~\ref{tab:seed_count}).
Despite a $4\times$ latency increase over the $5{,}000$-seed setting, we adopt
$10^5$ seeds as $\sim\!31$\,ms remains well within our control budget.

\begin{table}[!t]
\centering
\caption{Sensitivity to sampling stream (50 pseudo-random re-draws on 16
problems) against the deterministic Sobol set.}
\label{tab:seed_variation}
\begin{tabular}{@{}lcc@{}}
\toprule
Property & 10,000 seeds & 100,000 seeds \\
\midrule
Sobol pass rate & 93.75\% & 100.00\% \\
Pass rate (Mean $\pm$ SD) & $98.62 \pm 2.59$\% & $100.00 \pm 0.00$\% \\
Pass rate (Min / Max) & 93.75\% / 100.0\% & 100.0\% / 100.0\% \\
Median $t$ (Mean $\pm$ SD) & $8.5 \pm 1.9$ ms & $37.1 \pm 6.1$ ms \\
\bottomrule
\end{tabular}
\end{table}

At $10^5$ seeds all streams succeed on every problem
(Table~\ref{tab:seed_variation}), showing that the result does not depend on
the specific Sobol sequence. The residual spread at $10^4$ seeds confirms the
larger budget eliminates variance, not bias. Re-enabling the swivel stage
retains 100\% success at $\sim\!40$\,ms median latency, adding minimal
overhead to the isolated detour.

\section{Conclusion}
\label{sec:conclusion}

We posed obstacle avoidance under a prescribed trajectory as maximal task
preservation, classifying interference as reducible, absorbed by null-space
motion with no task deviation, or irreducible, where no configuration both
tracks the task and clears the obstacle. The classification is an emptiness
test rather than a threshold, so the task is modified only where it cannot be
executed. Where it must be, we search a roadmap of the configuration-space
obstacle boundary built in a corridor around the blocked segment, so the
direction of circumnavigation follows from comparison across that segment.
\textsc{Close\&Safe} attains the highest preservation of any collision-free
method across four simulated scenarios, and recovers more than $70\%$ of
two demonstrated strokes on hardware.

Preservation is spatial only. The repaired path is re-timed under a
minimum-jerk profile, so velocity and acceleration follow from the actuation
limits rather than from the demonstration, and extending the metric and the
repair to the velocity profile would close that gap. Two quantities remain
empirical: planning time is dominated by samples the boundary filter
discards, and the projection distance is fixed from trials. The obstacle set
is likewise static, though since the repair runs at
\SI{25}{\hertz}
moving obstacles are a question of re-evaluating feasibility as the scene is
updated rather than of planning cost.

\end{document}